\documentclass[conference]{IEEEtran}
\IEEEoverridecommandlockouts
\usepackage{cite}
\usepackage{amsmath,amssymb,amsfonts}
\usepackage{algorithmic}
\usepackage{graphicx}
\usepackage{textcomp}
\usepackage{xcolor}
\usepackage{booktabs}
\usepackage{hyperref}
\def\BibTeX{{\rm B\kern-.05em{\sc i\kern-.025em b}\kern-.08em
    T\kern-.1667em\lower.7ex\hbox{E}\kern-.125emX}}

\newcommand{\elt}[0]{\ensuremath{\ell_{\text{true}}}}
\newcommand{\eli}[1]{\ensuremath{\ell_{#1}}}
\newcommand{\elri}[2]{\ensuremath{(\eli{#1})_{#2}}}

\newcommand{\qli}[1]{\ensuremath{Q_{\ell_{#1}}}}
\newcommand{\rlt}[1]{\ensuremath{R_{#1,\elt}}}
\newcommand{\rlit}[2]{\ensuremath{R_{(#1)_{#2}; \elt}}}
\newcommand{\rlri}[2]{\ensuremath{R_{(\ell_{#1})_{#2}}}}
\newcommand{\summrunexample}{\ensuremath{(R_{a_i}=4, R_{b_i}=6)}}

\newcommand{\scsumm}{\ensuremath{(\rlri{1}{i}, \rlri{2}{i}, \ldots, \rlri{R}{i})}}

\begin{document}

\title{Logical Consistency Between Disagreeing Experts And Its Role In AI Safety
}

\author{\IEEEauthorblockN{Andr\'es Corrada-Emmanuel}
\IEEEauthorblockA{\textit{Data Engines} \\
andres.corrada@dataengines.com}}

\maketitle

\begin{abstract}
If two experts disagree on a test, we may conclude both cannot be
100\% correct. But if they completely agree, no possible evaluation can
be excluded. This asymmetry in the utility of agreements versus disagreements
is explored here by formalizing a logic of unsupervised evaluation for
classifiers. Its core problem is computing the set of group evaluations
that are logically consistent with how we observe them agreeing
and disagreeing in their decisions. Statistical summaries of their
aligned decisions are inputs into a Linear
Programming problem in the integer space of possible correct or incorrect
responses given true labels. Obvious logical
constraints, such as, ``The number of correct responses cannot exceed the number of observed responses," are inequalities. But in addition, there are ``axioms"—universally applicable linear equalities that apply to all finite tests. The practical and immediate utility of this approach to unsupervised evaluation using only logical consistency is demonstrated by building no-knowledge alarms that can detect when one or more LLMs-as-Judges are
violating a minimum grading threshold specified by the user.
\end{abstract}


\section{Introduction}

AI systems did not create  the principal/agent problem \cite{principalagentWiki}
but they are exacerbating it. We have an economic and
epistemological problem when we delegate to agents tasks that we do not want
to do, or cannot do because of our ignorance. Thus, we would like
to minimize our supervision while still remaining safe when we delegate
our decisions. Here we will consider how logical consistency between
possibly disagreeing agents can be used to ameliorate this ancient
problem (e.g. Plato's Allegory of the Ship of Fools in \emph{The Republic})
that AI systems cannot escape.

LLMs-as-Judges \cite{zheng2023judging} has emerged as one obvious way to
handle the problem of monitoring assistant LLMs when we do not
have access to a ground truth for evaluating their decisions. This,
however, does not resolve the recursive nature of the problem of
evaluating experts without answer keys to the tests we give them.
If we build a monitoring system, what or who monitors that? This is
the problem of \emph{infinite monitoring chains}—who judges the
judges?

No intelligence, however advanced, can escape the laws of logic
or physics \cite{tegmark2023}. We can use logical consistency
when evaluating experts without a ground truth for their
decisions (unsupervised evaluation). Tools built using logical
consistency alone will be, by construction, universally applicable
to any domain.

Using consistency alone, we can construct a logic that excludes
group evaluations that contradict observed statistical summaries
of expert responses.
The one-bit summary of two
experts in the abstract is a simple example of how the logic works.
It is also an example of how the exclusion of evaluations can serve
as the basis for no-knowledge alarms for classifiers.

The one-bit summary of the responses by the two experts carries
no information about the domain of the test. Nor do we have
any text or other semantic information to guide our evaluation.
Depending on its value, we can exclude a joint evaluation for
them (both cannot be 100\% correct). If our safety standard
had been that both classifiers had to be perfect in their
classifications, observing disagreement would have triggered
an alarm.

The logic will treat the classifiers as black-boxes
and only use summaries of how they decided or responded
when asked to classify items into a finite set of $R$
labels to be denoted by $\mathcal{L}.$ Such a logic cannot
be used in evaluating experts that go beyond
answering factual questions with a finite number of
answer choices \cite{kalai_why_2025}. But it can be used to evaluate the
judges of those experts. When we ask LLMs to act
as judges of the complex outputs of other LLMs
it often happens that we are asking them to be
classifiers. Graders are often classifiers.

Consider the MT-Bench \cite{zheng2023judging} dataset created to establish benchmarks
for evaluating LLMs-as-Judges when they carry out pair
comparisons between the complex outputs of LLMs. Each LLM-as-a-Judge
is presented with the outputs of two models that can be generically
denoted \emph{model\_a} and \emph{model\_b}. The judge must decide
which output they prefer or if both are equally good. In effect,
the LLMs-as-Judges are acting as classifiers by outputting one
of three possible grading labels—a, b, or tie.

This concrete example of the appearance of the classification task
when we consider the ability of LLM-as-Judges should make clear
that any logic of unsupervised evaluation for classifiers could
have wide applicability. Not as a way to evaluate agents carrying
out the primary task of the system, but as a way to evaluate agents
that are grading those primary ones.

Logical consistency is the only thing that can be established in
unsupervised settings where no ground truth is available. There
are no truth tables to allow us to formally prove that the decisions
of experts are correct. But we can prove or disprove the logical
consistency of various assertions we may want to make about tests,
their answer keys, and how well experts did answering them.

\subsection{Suspecting the answer key and all test takers}

Holistic evaluations of LLMs seek to measure their performance
across different dimensions \cite{liang2023}. Aligning these
systems to human preferences requires that we define a possible
ground truth for what that preference is. But human experts
are widely observed to never completely agree on that truth.
This occurs in the MT-Bench benchmark where 5K+ human judgments
show an agreement rate of 80\%. This may be thought to occur
because the human experts are judging on properties like "roleplaying"
ability. But agreement rates meaningfully below 100\% can also be
observed across doctors in different countries reviewing medical
cases \cite{kohane_isaac_2025}.

If we want to design AI systems that are human-centered,
what does that mean when human experts cannot agree on what
the answer key for a test is? This problem will be 
considered in section \ref{sec:suspecting} where the logic will be used to
establish an upper bound on the minimum accuracy of human
experts given that we believe there is a correct answer
key to the test they took.

\section{The basic question in logics of unsupervised evaluation}

We are discussing a logic of unsupervised evaluation for classifiers.
But it is instructive to consider what properties we would want from
anything that claimed to be a logic of unsupervised evaluation. Classification
is just one of the many tasks ML systems carry out. Accordingly, we would
expect there to be a logic of unsupervised evaluation for regressors, etc.

In each case, a basic question for such a logic must be—what are the
group evaluations logically consistent with how we observe test takers
agreeing/disagreeing in their decisions? For each case, we can build
statistical summaries of the events of the disagreement. For one dimensional
regressors we would have the histogram of their pairwise differences, for
example. And for classifiers we can summarize the question-aligned responses
as the integer counts of the $R^N$ ways that $N$ classifiers can
agree/disagree when doing $R$-label classification.

We can formalize this further for classifiers by considering the
trivial ensemble, $M=1$. In this case, the basic question can
be stated as,
\begin{quote}
    Given the summary of classifier $i$'s test responses as
    label integer counts, $(R_{\ell_1},R_{\ell_2},\ldots,R_{\ell_R})_i$,
    what are the possible integer points,
    \begin{multline}
    \label{eq:single-classifier-eval-space}
    (Q_{\ell_1}, Q_{\ell_2}, \ldots, Q_{\ell_R}) \; \otimes \\ \prod_{\elt \in \mathcal{L}} 
    (R_{\ell_1; \elt}, R_{\ell_2; \elt}, \ldots, R_{\ell_R; \elt}),
    \end{multline}
    logically consistent with those observations?
\end{quote}

This integer space for enumerating the possible evaluations of
a single classifier is composed of two types of subspaces. First
we have the space for unknown statistics of the answer key for
the test. For $R$-label classification, that space is of dimension
$R.$ But note that not all points in this space are consistent
with observations in a test of size $Q$. In particular, any
properly constructed answer key projects to non-negative integers
that satisfy the equality,
\begin{equation}
    \sum_{\ell_r \in \mathcal{L}} \qli{r} = Q
\end{equation}
If there is an answer key to the test, its label counts
project to a point,
$(\qli{1}, \qli{2}, \ldots, \qli{R})$, in what we will
call the $Q$-complex (QC).
And once we are at that point in the QC, all the possible number of responses
, $R_{\ell_r; \elt}$, of $\ell_r$ given true label, \elt, must be on the
plane defined by,
\begin{equation}
    \sum_{\eli{r} \in \elt} \rlt{\eli{r}} = \qli{\text{true}}.
\end{equation}

All logical statements about evaluations can only occur at a single
fixed point in the $Q$-complex. When we are using how test takers
agree and disagree on a given test, there is no other possibility.
This will be used when we discuss how to construct the alarm comparing
classifier performance at the same assumed point in the QC.

Having picked a point in the QC, we can also lower the
dimensionality for the subspaces associated with 
possible correct responses given true label. These
are the spaces associated with tuples of the form,
\begin{equation}
    \{(R_{\ell_1; \elt}, R_{\ell_2; \elt}, \ldots, R_{\ell_R; \elt})\}_{\elt \in \mathcal{L}}.
\end{equation}
For these spaces we require,
\begin{equation}
\{\sum_{\eli{r} \in \mathcal{L}} R_{\eli{r}, \elt} == \qli{\elt}\}_{\elt \in \mathcal{L}}
\end{equation}

\begin{figure}[htbp]
\centerline{\includegraphics[width=\columnwidth]{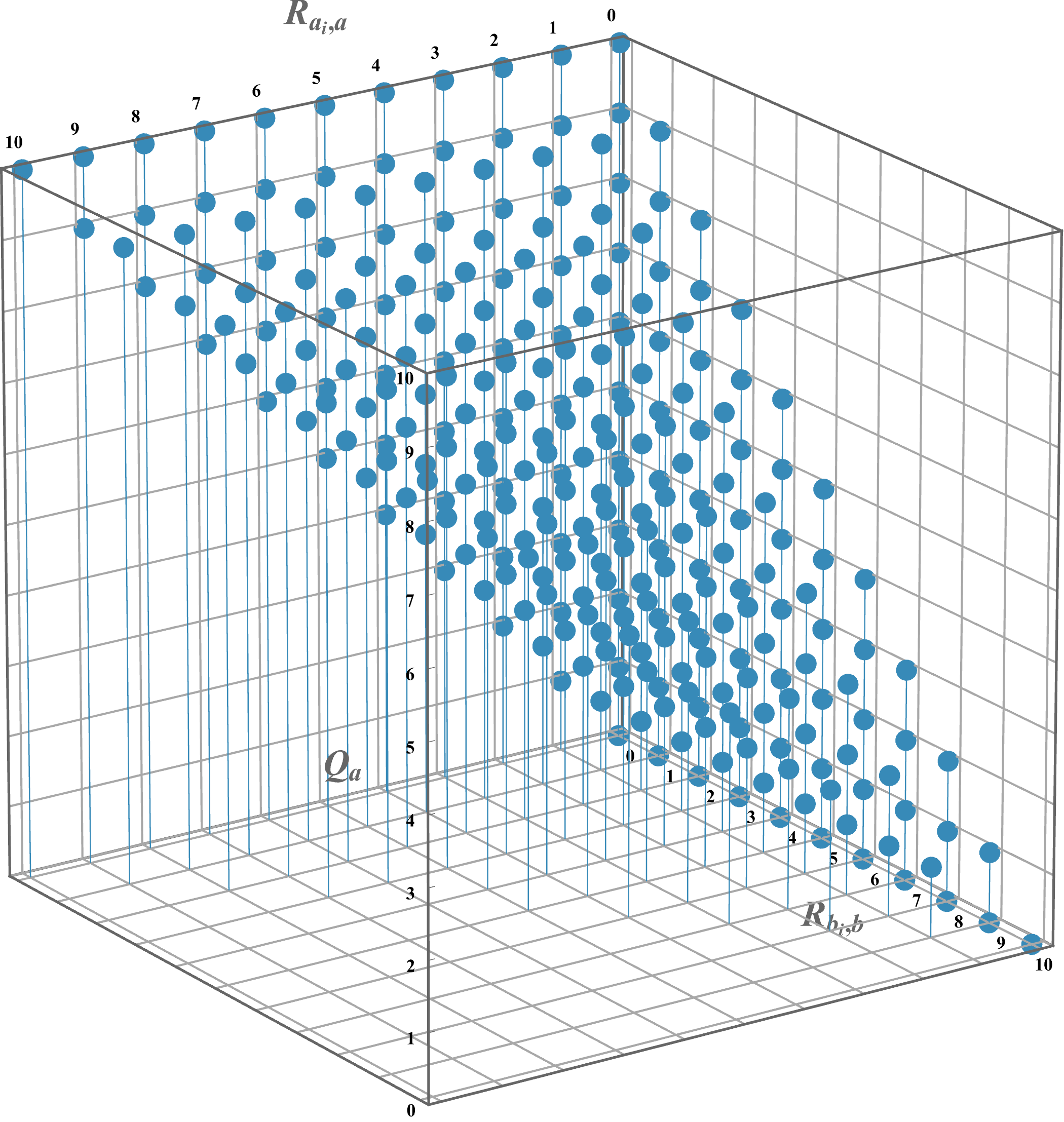}}
\caption{All possible evaluations for a binary classifier labeling $Q=10$ items.
$R_{a_i,a}$ and $R_{b_i,b}$ are the number of correct responses by classifier
$i$ for labels `a' and `b' respectively.}
\label{fig:all-possible}
\end{figure}

As a consequence of these plane equalities, possible evaluations
exist within an $(R-1)(R+1)$ dimensional space within the $R(R+1)$
space in (\ref{eq:single-classifier-eval-space}).
For binary classification ($R=2$) this is a 3-dimensional
space shown in Fig.~\ref{fig:all-possible}.
Note that the possible set does not fill
the space and has non-trivial geometry.

We could do all of the logic in a space more familiar to
the ML community. We could represent a binary classifiers
evaluations as points in the space defined by,
\begin{equation}
    (P_{a_i,a} = \frac{R_{a_i; a}}{Q_a}, P_{b_i,b} = \frac{R_{b_i; b}}{Q-Q_a}, P_a = \frac{Q_a}{Q}).
\end{equation}
The possible set of evaluations in that space for a test of size
$Q=10$, the same as Figure 1, is shown in Figure 2. It is clear that
what we will call the $R$-space representation is easier to
visualize than the $P$-space typically used in ML metrics. The two
representations have their own use, however. For example,
completeness of the axioms for classifiers is easier to prove in
$P$-space as discussed in the Appendix. The rest of the paper
will continue in the $R$-space representation but equivalent
constructions are possible in $P$-space.

\begin{figure}[htbp]
\centerline{\includegraphics[width=\columnwidth]{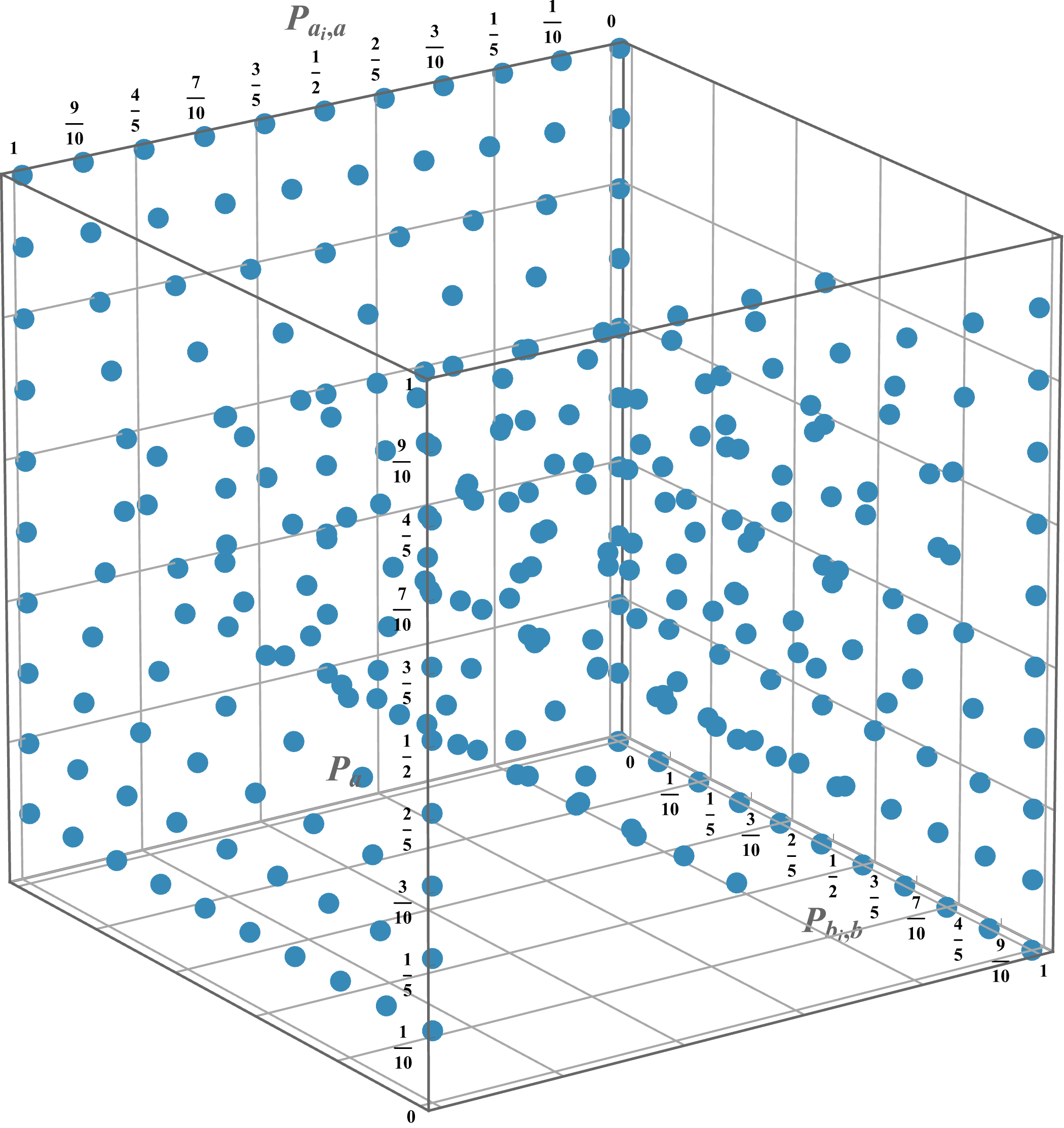}}
\caption{All possible evaluations for a $Q=10$ test but now in the
space of prevalence,$P_a=Q_a/Q$, and label accuracies, 
$P_{a_i,a}=R_{a_i,a}/Q_a$ and $P_{b_i,b}=R_{b_i,b}/(Q-Q_a).$
This is visual proof that the geometry of possible evaluations
is easier in the integer response space shown in Fig.~\ref{fig:all-possible}.}
\label{fig:pspace-all-possible}
\end{figure}

The set of all possible evaluations before we observe a test
taker $i$'s responses can be reduced further
once we observe a summary of their responses as 
$(\rlri{1}{i}, \rlri{2}{i}, \ldots, \rlri{R}{i}).$ Then,
it must be true for any label, \elt, that,
\begin{equation}
\label{eq:inequality-constraints}
\rlit{\eli{r}}{i} \leq \rlri{r}{i}.
\end{equation}
\begin{figure}[htbp]
\centerline{\includegraphics[width=\columnwidth]{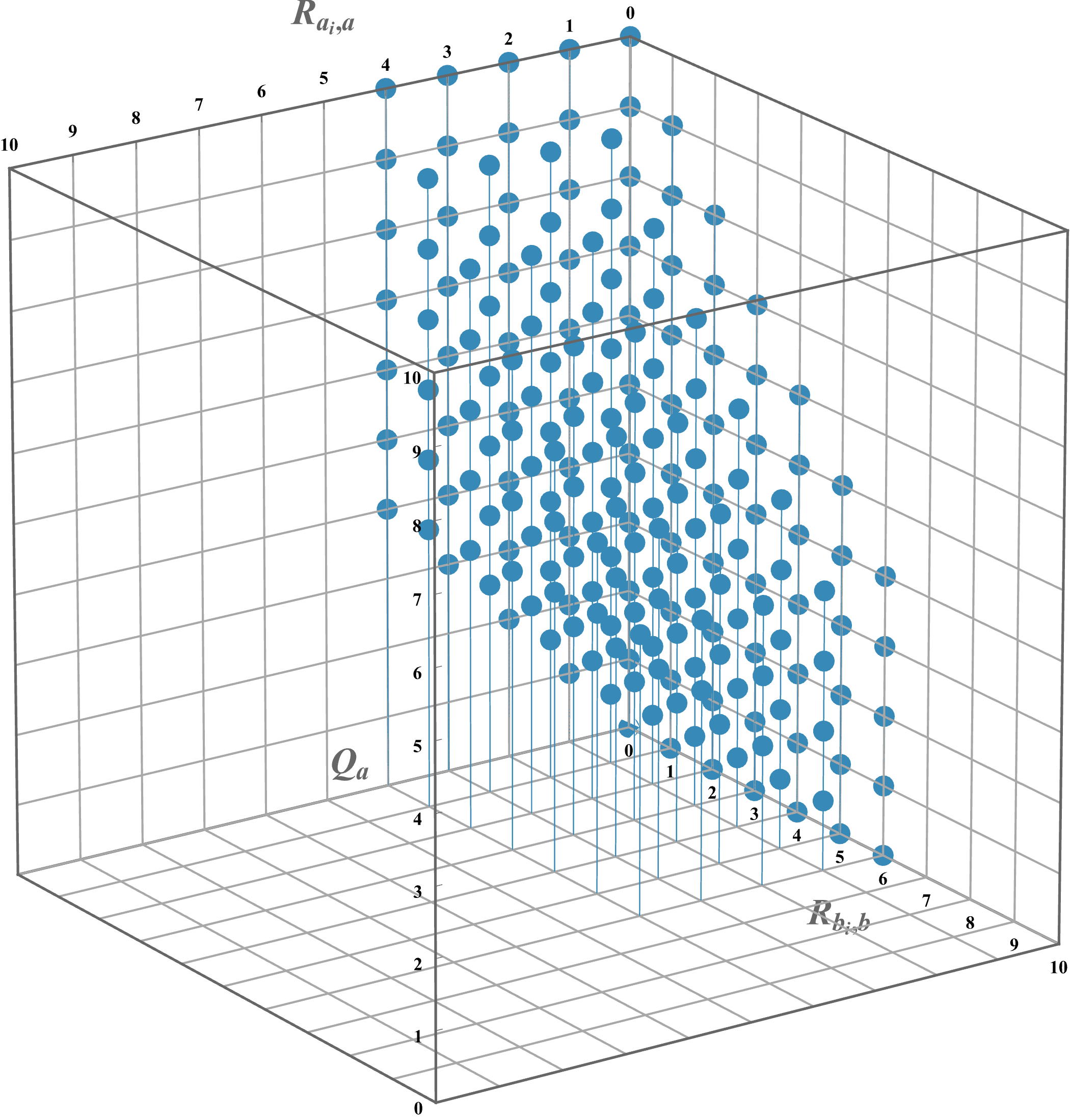}}
\caption{All possible evaluations for a $Q=10$ test after we observe the
test summary $(R_{a_i}=4, R_{b_i}=6).$ The inequalities change the number
of possible evaluations but not the dimension of their geometry.}
\label{fig:after-inequalities}
\end{figure}

In Fig.~\ref{fig:after-inequalities} we see the effect of these 
inequality constraints
for the binary classification example where we have observed
$(R_{a_i} = 4, R_{b_i} = 6)$ and can thus conclude that the
only possible evaluations must obey,
\begin{equation}
R_{a_i;a} \leq 4, \; R_{b_i; b} \leq 6.
\end{equation}
In this $Q=10$ example, there are 285 possible evaluations
and the inequalities after having summarized the test responses
reduce this to 235 but do not change the dimension of the
possible set given test results. To do that, one must
discuss a set of axioms that relate label response counts
across the true labels.

\subsection{Axioms for the single classifier}

\begin{figure}[htbp]
\centerline{\includegraphics[width=\columnwidth]{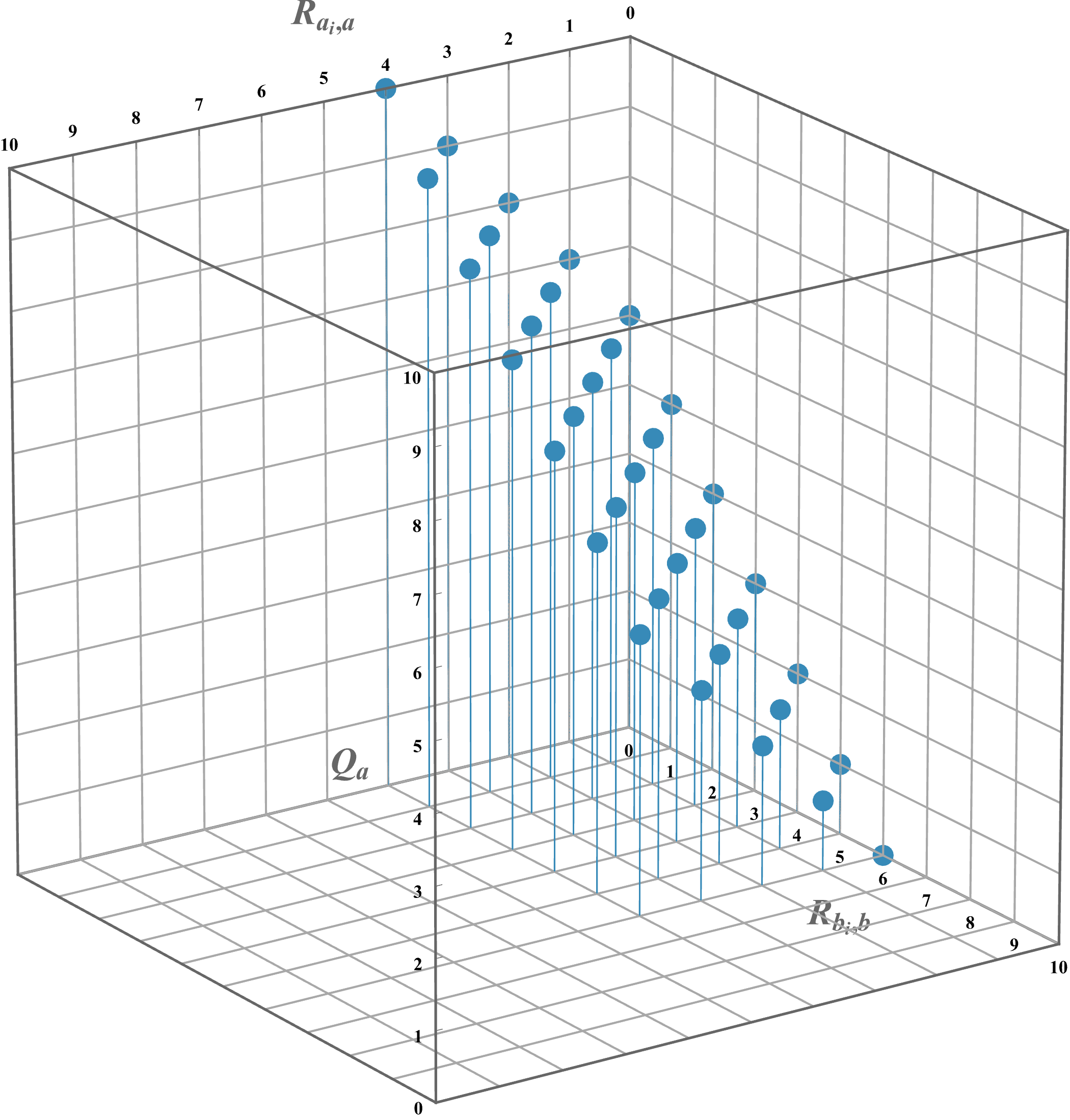}}
\caption{All possible evaluations for a $Q=10$ binary test after we observe the
test summary $(R_{a_i}=4, R_{b_i}=6)$ and pick evaluations consistent
with it as expressed by either of the binary axioms: $R_{a_i,a}-Q_a + (R_{b_i}=6) - R_{b_i,b},$
or $R_{b_i,b}-Q_b + (R_{a_i}=4) - R_{a_i,a}.$}
\label{fig:after-axiom}
\end{figure}

We now discuss the set of equalities that must be
obeyed by any classifier $i$ taking a test of size $Q$
and where we observe they respond with
summary \scsumm. For each label in the set of $R$ labels,
called \elt, and each classifier $i$ in an ensemble,
the following linear relation is always equal to zero,
\begin{multline}
    \label{eq:single-axioms}
    \rlit{\elt}{i} - \qli{\text{true}} + \sum_{(\eli{r} \neq \elt) \in \mathcal{L}} R_{(\eli{r})_i}\\
    - \sum_{(\ell \neq \elt) \in \mathcal{L}} \sum_{(\eli{r} \neq \elt)} R_{(\eli{r})_i; \ell}.
\end{multline}
There are $R$ of these equations, one for each
label, as we would expect from symmetry when no ground truth is available. It cannot be
the case that just observing decisions can privilege any one label. Their simple proof
is given in the Appendix. They reduce the size of the possible set and also the dimension
of its geometry. Fig.~\ref{fig:after-axiom} shows
the effect of the axioms for the $Q=10$ binary test example.

It is instructive to work through an example of a possible
evaluation after the inequality constraints 
\eqref{eq:inequality-constraints} are imposed that is
impossible given the observed counts.
Consider the running $Q=10$ example with observed responses
\summrunexample. A possible evaluation is 
$(R_{a_i,a} = 3, R_{b_i,b} = 1, Q_a = 7).$ It satisfies the
inequality constraints \eqref{eq:inequality-constraints}, but
cannot be correct given the observed test responses.

If they have only
one label `b' decision correct ($R_{b_i,b} = 1$), that means
5 of their `b' decisions were actually `a's. But there are
only 7 `a's in the answer key, so the classifier cannot
be 3 times correct on `a' decisions ($R_{a_i,a} = 3$). It
can only be $R_{a_i,a}=2$ everything else being equal. The
evaluation $(R_{a_i,a} = 2, R_{b_i,b} = 1, Q_a = 7)$
obeys the axioms \eqref{eq:single-axioms}.

The equations are complete (see Appendix) but not independent,
they sum to identically zero. So in binary classification, $R=2$,
we can define the set of possible evaluations as lying on a plane,
but in general they reduce the dimension by $R-1.$

The filtering effect of the axioms is stronger than that of the response inequalities
since they define reduce the dimension of the possible set.
In the running example of a single binary classifier labeling $Q=10$ items,
there are 286 possible evaluations. In general,
there would be evaluations $1/6 (Q+1)(Q+2)(Q+3)$ for a single binary classifiers
(Appendix). The inequality constraints \eqref{eq:inequality-constraints} 
reduce this to 210 and the axioms \eqref{eq:single-axioms} to 35.
Roughly, the single classifier choosing between $R$ labels is
going to have $Q^{R(R-1)}$ evaluations and this is reduced by $(R-1)$ dimensions
by these axioms. Thus, the percentage reduction in possible evaluations goes
as,
\begin{equation}
    \frac{1}{Q^{(R-1)}}.
\end{equation}

\subsection{Group evaluations logically consistent with $M=1$ summaries}

If we have the question-by-question responses of each test taker,
sets of the form $\{(q,\rlri{r}{i})\}$, we could proceed to consider 
summaries of
the test for any subset of size $M$ for the $N$ classifiers. So
far, we have been considering the $M=1$ summaries, \scsumm. And
we will continue to do so since this allows us to avoid the problem
of considering how classifiers are correlated in their decisions.

Note however, that the algebra of the logic allows us to conclude
that, whatever the set is that fully takes into account $M=m>1$
summaries, it must be contained within the product of the $m$ $M=1$
sets.
It cannot be the case that evaluations inconsistent with the $M=1$
summary of a classifier, become consistent given pair summaries
with any other classifier. Whatever that pair summary is, it has
to project to the single summary that defines the $M=1$ possible set.

One way to view this is as a ladder of logical inference. We first
have to assume a location point in the QC for the summary of the answer key.
The previous section then discussed the evaluations possible given
$M=1$ summaries. To obtain the evaluations logically consistent
with the $M=2$ summaries, we start from the product space of
the $M=1$ summaries and then consider what points are excluded
from that product space given the $M=2$ axioms.

\begin{figure}[htbp]
\centerline{\includegraphics[width=\columnwidth]{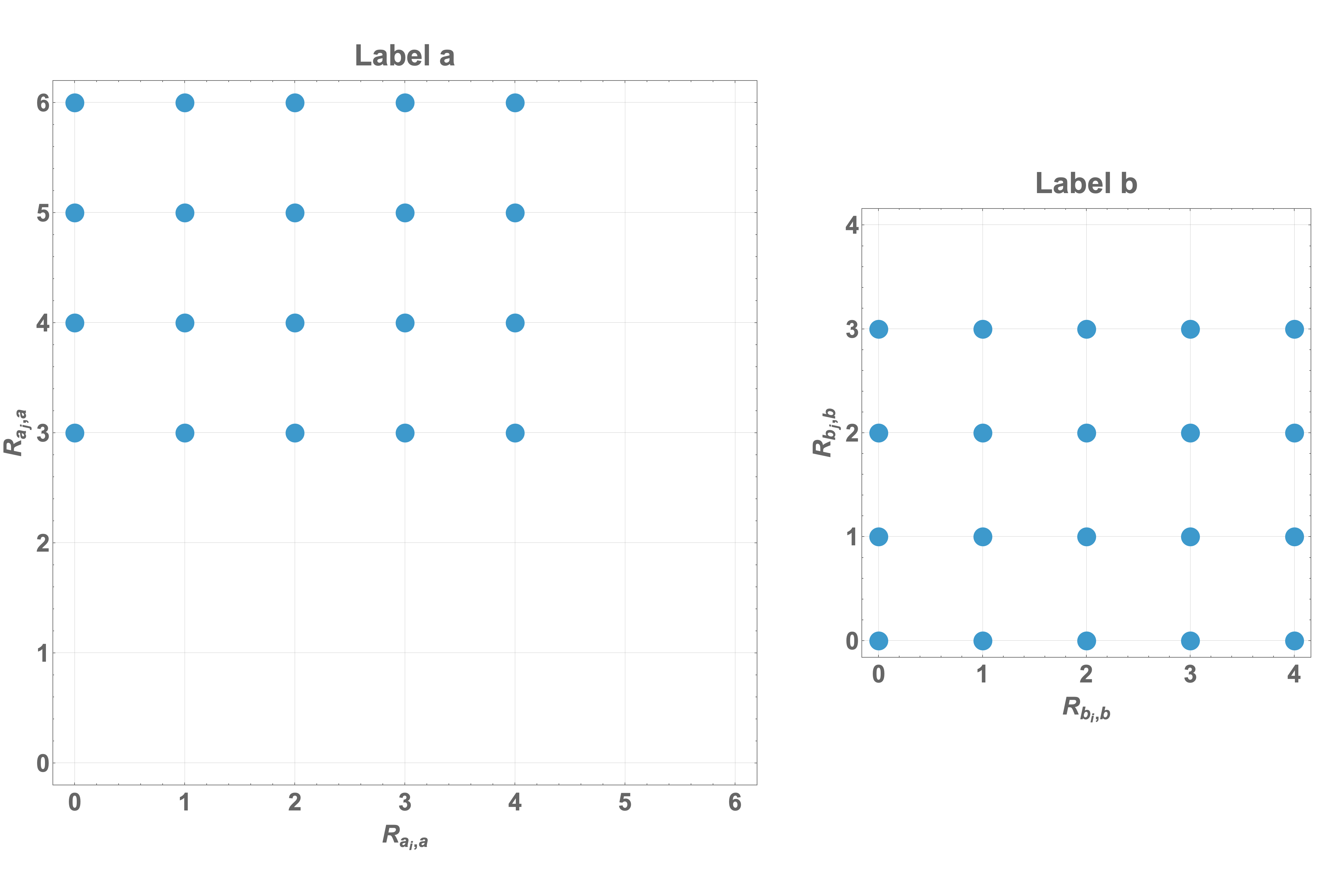}}
\caption{All possible evaluations for a $Q=10$ binary test assuming the answer
key summary is $(Q_a=6, Q_b=4)$ and we have observed the test summaries
$(R_{a_i}=4, R_{b_i}=6)$ and $(R_{a_j}=7, R_{b_j}=3)$ for classifiers
$i$ and $j.$}
\label{fig:binary-pair-product-space}
\end{figure}

We can visualize this product space at each point in the QC for a pair
of binary classifiers as two square spaces, one for each label as in,
for example, Fig.~\ref{fig:binary-pair-product-space}. These are
their possible evaluations at QC point $(Q_a=6,Q_b=4)$ consistent 
with $M=1$ axioms given
that we observed individual test responses $(R_{a_i,a}=4, R_{b_i,b}=6)$
and $(R_{a_j,a}=7, R_{b_j,b}=3).$ 
Beyond binary classification, this becomes harder or impossible
to visualize so we will use \emph{correctness squares or cubes}, 
where we plot the number of correct responses for each true label in
separate figures, and each label figure has axes variables corresponding
to the different classifiers. By construction, we are throwing away
the count of error decisions. So for $R>2$ classification, we can
include the multiplicity of evaluations for each possible correct point.
The next section will discuss this as we proceed to explain the atomic
logic operation for no-knowledge alarms of misaligned classifiers.

\section{No-knowledge alarms for misaligned classifiers}

The MT-Bench benchmark \cite{zheng2023judging} was created to
evaluate the performance of LLMs-as-Judges. The concept of using
other agents to monitor the work of primary agents is well-known
as a safety design pattern through-out history. But as work on
the principal/agent problem in economics shows \cite{laffont_theory_2002}, 
using supervisors to manage the work of others, does not resolve all monitoring
issues.

In particular, using AI agents to monitor others in unsupervised settings,
as LLMs-as-Judges are intended to do, still leaves unresolved who monitors
the judges. The logic being presented here based on consistency between
agreement/disagreement counts of test-takers can alleviate this unresolvable
problem. Absent ground truth for the decisions of experts and any other side
information, using other experts to grade them leads to 
\emph{infinite monitoring chains}.

But note that as the monitoring chain extends, the evaluation of the judges
becomes easier and simpler. This is the case with the core evaluation task
in the MT-Bench benchmark -- pair comparisons of the complex output of two
LLMs. The use of pair comparisons means that, in effect, the expert comparing
their outputs is acting as a 3-label classifier. Presented with "model\_a"
and "model\_b" outputs, a judge can pick either one (two labels) or the
third label ("models tied"). So we can use the logic developed here to
build alarms for such pair comparison agents.

\begin{figure}[htbp]
\centerline{\includegraphics[width=0.85\columnwidth]{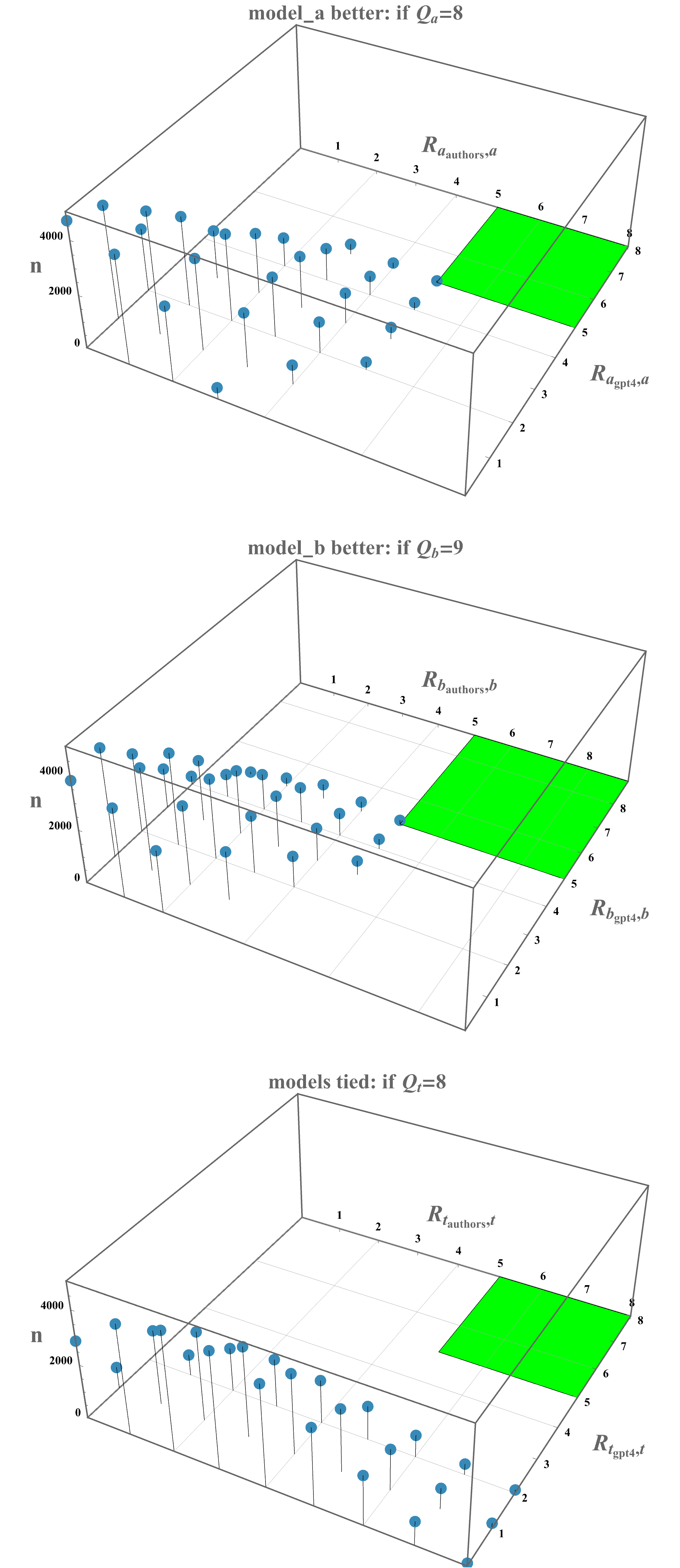}}
\caption{The multiplicity count of possible number of correct responses by 
two experts doing
pair comparisons of LLM outputs (model\_a and model\_b). They graded 25
pairs from the MT-Bench dataset. Their grade summaries are,
$(R_{a_i}=5,R_{b_i}=10,R_{t_i}=10)$ and $(R_{a_j}=4,R_{b_j}=18,R_{t_j}=3)$
for \texttt{authors} and \texttt{gpt4}, respectively.
There are no possible evaluations, given these summaries, and the assumption
that the correct answer key for the test has summary $(Q_a=8, Q_b=9, Q_t=8)$, 
that has both
classifiers grading the tied pair comparisons better than 25\%. The green
squares represent points were both classifiers are more than 50\% correct.}
\label{fig:mt-bench-correct}
\end{figure}

The basic idea of the alarms is that the possible set of classifier
evaluations at any point in the QC excludes some evaluations.
This means we can build no-knowledge alarms for evaluation conditions
that are label and classifier symmetric in wholly unsupervised settings.
For example, in Fig.~\ref{fig:mt-bench-correct}, the product space of $M=1$ evaluations
for two experts grading 25 pair-comparisons from the MT-Bench benchmark
are shown. These are the possible number of correct label responses for
two graders (the \texttt{authors} of the MT-Bench dataset and \texttt{gpt4})
that we
have observed to respond with summaries $(R_{a_i}=5,R_{b_i}=10,R_{t_i}=10)$
and 
$(R_{a_j}=4,R_{b_j}=18,R_{t_j}=3)$
for \texttt{authors} and \texttt{gpt4}, respectively.

The \texttt{authors} decision is the weighted vote of the authors
of the MT-Bench dataset \cite{zheng2023judging}. \texttt{gpt4}
is the LLM-as-Judge grader of the pair comparisons of two LLM outputs.
Details about the construction of the voted decisions for the \texttt{authors}
and the construction of a ground truth decision using human expert
judgments are in Appendix \ref{sec:app-gt-construction}. We see in
Fig.~\ref{fig:mt-bench-correct} that there is no evaluation for
\texttt{gpt4} where it is doing better than about 25\% correct
on the pair comparisons that human experts deemed a tie.

If we had set the alarm to trigger on any label performance
being less than, say, 50\% correct, the observed test summaries for
\texttt{authors} and \texttt{gpt4} would have triggered it
for this particular $Q=25$ test. We have verified that at this
QC point, there is at least one classifier violating the threshold
for one of the labels.
But this observed failure occurred at a single point in the
QC. For a test of size $Q=25$ with $R=3$ possible responses,
there are 351 points of the form $(Q_a, Q_b, Q_t)$ in the
QC. It may be that at some other assumed summary of the answer
key, both graders are better than 50\% or some other value
for both labels. It so happens that for this particular
$Q=25$ test a trigger
set to any value greater 46\% would be able to detect one of the classifiers
is misaligned with human experts on one of the labels.

\begin{figure}[htbp]
\centerline{\includegraphics[width=\columnwidth]{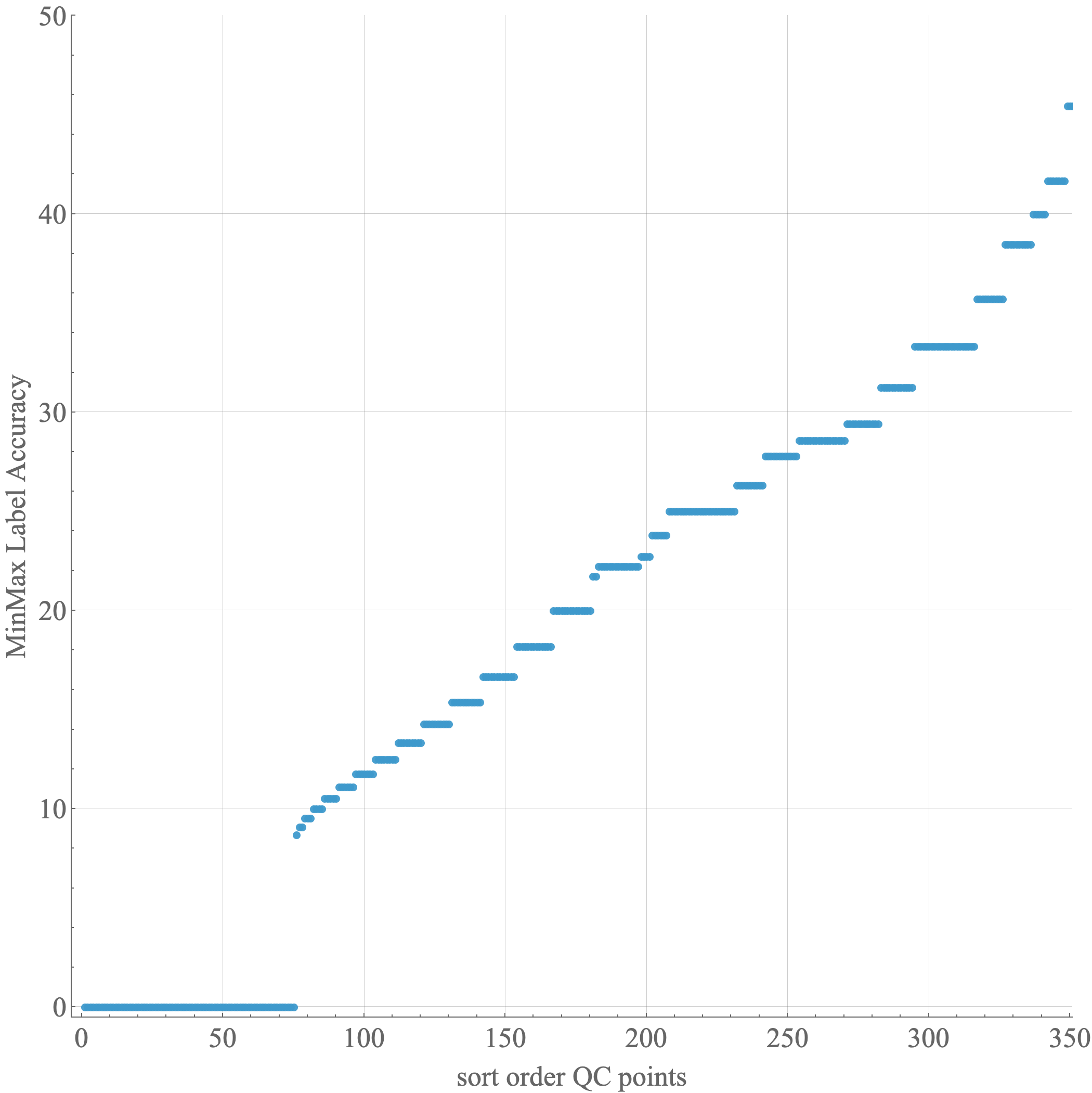}}
\caption{Plot of the minimum value of the max label accuracy
for both graders of the $Q=25$ MT-Bench pair comparison test.
For the observed grade summaries,
$(R_{a_i}=5,R_{b_i}=10,R_{t_i}=10)$ and $(R_{a_j}=4,R_{b_j}=18,R_{t_j}=3)$,
for \texttt{authors} and \texttt{gpt4} respectively, any
alarm condition set at more than 46\% would trigger. The 351
points in the QC of a 3-label $Q=25$ test have been ordered
by the min-max value.}
\label{fig:mt-bench-trigger}
\end{figure}

We can determine, universally, what minimum threshold value
for label accuracy would be triggered by the observed
single classifier summaries. This can be visualized as
a plot of the minimum of the maximum label accuracies
for both classifiers as shown in Fig.~\ref{fig:mt-bench-trigger}.
The max of this min-max plot is at about 45\%. Thus any setting greater than
this could be used to trigger a monitoring alarm that would have
gone off with these summaries. It is worthwhile noting that
the worst performing accuracy is actually about 14\% on
the tied judgments of gpt4. Logical consistency can only upper
bound the accuracy that would be triggered given the observed
$M=1$ summaries, in this case.

\begin{figure}[htbp]
\centerline{\includegraphics[width=\columnwidth]{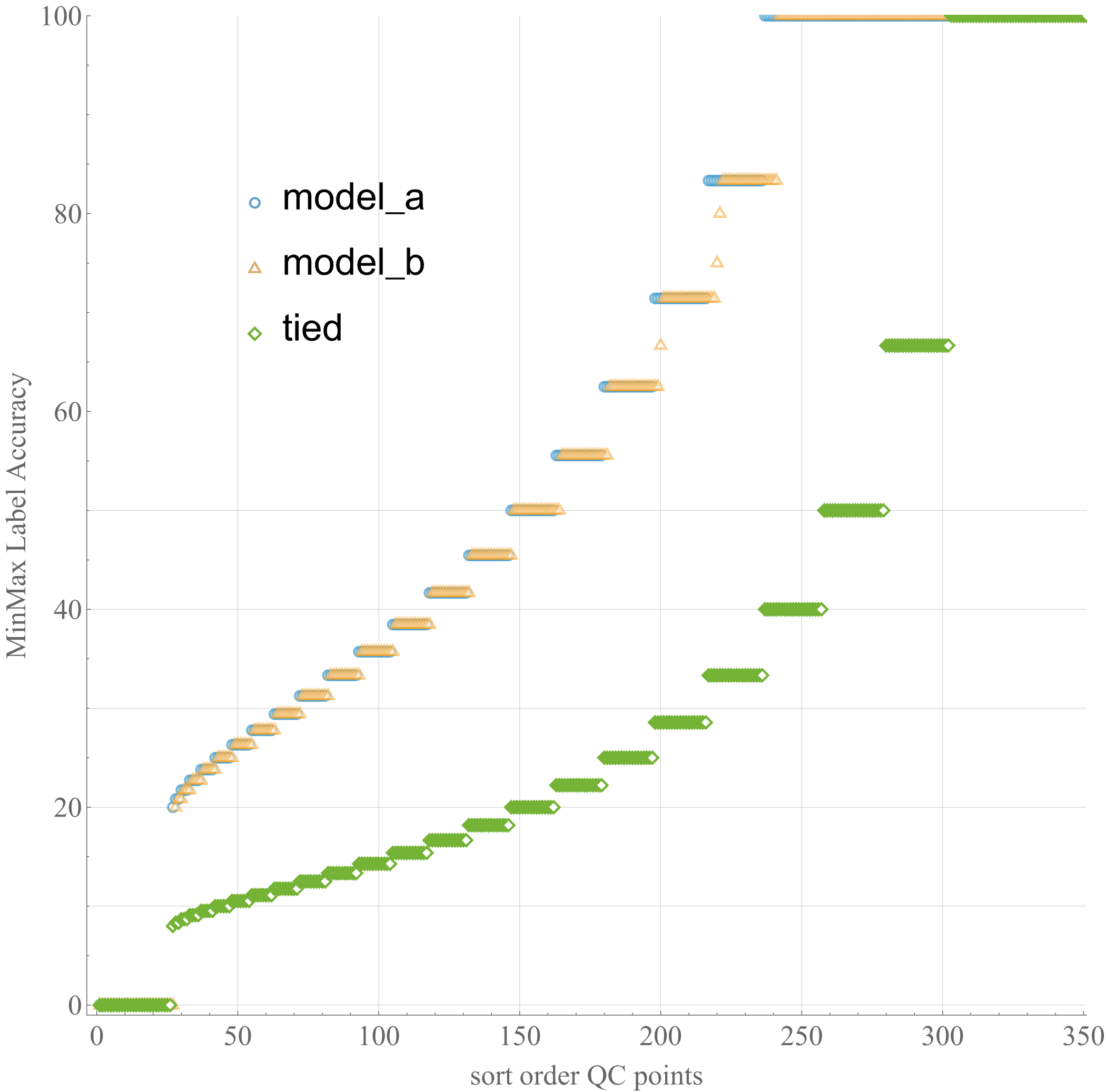}}
\caption{Plot of the minimum value of the max label accuracy, by label,
for both graders of the $Q=25$ MT-Bench pair comparison test.
We cannot set an alarm on an individual label since, for any label,
there are answer summaries for any accuracy level.The 351
points in the QC of a 3-label $Q=25$ test have been order
by the min-max value, per label. So points at this plot
cannot be compared across label for fixed sort order.}
\label{fig:mt-bench-by-label-trigger}
\end{figure}

Note that this alarm is predicated on an accuracy threshold that
is label and classifier independent. There is no privileged
label or classifier when logical consistency is all we are using.
If we plot the min-max by label -- the mininmum value of the maximum
possible accuracy for both classifiers -- as shown in Fig.~8, we
see that no threshold is possible by label. For any label, there
will always be answer key summaries that make it impossible to
exclude any possible evaluation from 0\% to 100\%. We do not
discuss the setting of alarms that set different thresholds
per label but require the conditions to hold simultaneously
for all labels at any point in the QC.

\section{Suspecting the ground truth}
\label{sec:suspecting}
Logical consistency is not magical. It cannot sense the presence of unknown
answer keys. Strictly speaking, the no-knowledge alarms are not establishing
that classifiers are misaligned to an unknown ground truth. They are establishing
that there is no answer key, given the count of test responses, that has
all classifiers obeying the accuracy specification at the set trigger level.

It could be that there is no answer key for the test we have formulated.
Logical consistency cannot answer those scientific questions. Those depend
on knowledge of the domain where the logic is applied. The logic is
universally applicable to any domain, but it cannot establish whether the
tests we use are correct or not.

There is a practical way to use this inability of what assumption in
our statements is the false one. Consider the widespread use of
agreement rates between human experts to establish the answer key
for evaluations. This is the case with the MT-Bench dataset we have
been using. It contains about five thousand \emph{human expert}
pair comparison judgments across different dimensions of LLM performance.

The agreement rate between the MT-Bench human experts is about 80\%
\cite{zheng2023judging}. This is about typical with other agreement
rates, even in medical contexts where we may expect the truth to
be easier to ascertain by experts. So how come we believe that we
can align AI systems to human values when humans, even those
considered experts in a domain, are observed to disagree at rates
of 20\% or higher \cite{salaudeen2025}?

The approach presented here suggests that we should "invert" the
process of discussing ground truths established by disagreeing
experts. We can ask, what is the trigger threshold set by the
observed disagreements on a supposedly existing ground truth?
Whether human experts disagreeing 20\% on a ground truth for
evaluating LLM outputs is a cause for concern or not is not
resolved by just being told that they agree 80\%. That sounds
good enough. Is it? This formalism turns the observed counts
of disagreements into a statements of the form,
\begin{quote}
    The ground truth established by these $N$ experts is
    logically consistent with them being x\% or more
    correct on all the labels.
\end{quote}
We argue, in fact, that this should become a minimum standard
for scientific work that reports answer keys established by
disagreeing experts. The statistics of how they disagree and
what this implies for their minimum competency based
on the belief of the existence of a correct, but unknown, answer key
can be answered by this logic. And can serve as a surrogate for
the "error" in the answer key constructed by ensembles of imperfect
classifiers.

\section{Conclusions}

Disagreements between graders of 25 pair comparisons of LLM
outputs from the MT-Bench benchmark have been shown to be
sufficient to trigger alarms of the form,
\begin{multline}
    ((P_{a_i,a} > x\%) \land (P_{a_j,a} > x\%)) \land \\
    ((P_{b_i,b} > x\%) \land (P_{b_j,b} > x\%)) \land \\
    ((P_{t_i,t} > x\%) \land (P_{t_j,t} > x\%)),
\end{multline}
for any $x$ greater than 46\%. This was done by looking
at the set of possible evaluations logically consistent
with the $M=1$ summaries, \scsumm, as established by
the single axioms \eqref{eq:single-axioms} as applied
to three label classification.

The limitations of logical consistency alone should be
clear. It can never detect that classifiers agreeing in their
answers are incorrect. Most importantly, these logical
considerations cannot establish the scientific validity
or usefulness of using these alarms in any context. Those
remain scientific and engineering questions that no logic
could universally answer.

These no-knowledge alarms should be viewed as simple devices
that can form part of a greater monitoring system. Like smoke
alarms, they cannot determine what is causing the problem or
how to fix it. But knowing that something is definitely wrong,
with logical certainty, can be used to increase the safety of
actions taken using the decisions of disagreeing experts,
whether human or robotic.

\section{Related Work}

Unsupervised evaluation has been treated in the ML literature.
Most work has focused on making point estimates of possible
correctness of experts, not its logic and what consistency
implies for the possible set of evaluations given test response
summaries.
Dawid and Skeene \cite{Dawid79} first treated the problem of
grading medical doctors without the ground truth of their
case decisions. Bayesian approaches were started by Raykar
\cite{Raykar2010} and further developed by Xang, Xi, Zhung,
and Jordan \cite{Zhang2014}. Parisi and subsequently Nadler,
have developed a spectral approach \cite{Parisi1253,Jaffe2015, Jaffe2016}.

As noted, all these approach are probabilistic and did not
discuss the concept of possible evaluations being restricted
by universally applicable axioms as is being claimed here. This
paper, however, is not the first to discuss logical aspects
of unsupervised evaluation. The first proof of unsupervised
axioms was done by Platanios and his \emph{agreement equations}
\cite{Platanios2014, Platanios2016}. It should be clear from this
paper that agreements are not sufficient nor very useful as
ways to keep track of classifier performance. The agreement
equations are part of the polynomials we can write for all
the possible ways $N$ classifiers can agree/disagree. There
are $R^N$ of these polynomials, only $R$ of them are about
the agreement events for all of them.

To fully understand the axioms for any $M=m$ summary, one must
go to the P-space we studiously avoided in the paper. In that
space it becomes possible to quickly write polynomials of
the unknown evaluation values (prevalences, label accuracies,
and their error correlations). The tools of algebraic geometry
can then be used to establish factorizations of those polynomials.
The use of algebraic geometry in statistics was pioneered by
Pistone \cite{pistone_algebraic_2000}.

\bibliographystyle{IEEEtran}
\bibliography{references}

\appendices
\section{Proof of the $M=1$ Axioms}
\label{sec:app-proof}
The proof the $M=1$ axioms for any number of labels is straightforward and
proceeds by expressing all variables in the axiom in terms of 
responses by true label.
For example, for any observed count of \eli{r} by classifier $i$, \rlri{r}{i},
it must be true that,
\begin{equation}
    \rlri{r}{i} = \sum_{\elt \in \mathcal{L}} R_{\elri{r}{i}, \elt}.
\end{equation}

Likewise, for any \qli{\text{true}}, we can express it in terms of
the classification correct and incorrect decisions as,
\begin{equation}
    \qli{\text{true}} = \sum_{\eli{r} \in \mathcal{L}} R_{\elri{r}{i}, \elt}.
\end{equation}
The axioms then follow trivially by linear cancellation of all terms.

\section{Constructing the ground truth and expert decisions}
\label{sec:app-gt-construction}

The majority vote of human experts in the MT-Bench pair comparisons was done by
weighted voting. An expert vote was split into two $1/2$ votes in the case
of ties in a pair comparison, but kept as $1$ if model a or model b was picked.
Thus, for example, if two experts voted (model a, tie) this would be consider
a preference for model a. Since we are using the tie grading label, this procedure
is unambiguous in assigning one of the three available grades.
The same procedure was used to establish the grades by the panel of \texttt{authors}.
In that case, we allowed pair comparisons that had just grades by one author. The
turn 1 pair comparisons used are show in Table~\ref{human-ground-truth}.

As discussed in the paper, logical consistency cannot establish that this construction
of \emph{a ground truth} is scientifically valid or useful. This MT-Bench example is
meant to illustrate the algebraic operations of the logic and that differences can trigger no-knowledge alarms
set to practical thresholds of safety.

\begin{table*}[htbp]
\label{human-ground-truth}
\centering
\caption{Human expert weighted vote ground truth, \texttt{experts},for the 25 LLM pair
comparisons from the MT-Bench dataset. There are 4 model\_a grades, 14
model\_b, and 7 ties.}
\begin{tabular}{c c c c c c}
\toprule
Question id & Model A & Model B & \texttt{experts} & \texttt{authors} & \texttt{gpt4}\\
\midrule
82  & gpt-4            & claude-v1        & b & a & a\\
82  & llama-13b        & gpt-4            & b & b & b\\
85  & vicuna-13b-v1.2  & gpt-3.5-turbo    & tie & tie & b\\
91  & gpt-3.5-turbo    & gpt-4            & tie & tie & a\\
91  & llama-13b        & gpt-3.5-turbo    & b  & b & a\\
91  & vicuna-13b-v1.2  & gpt-3.5-turbo    & a & tie & b\\
98  & vicuna-13b-v1.2  & gpt-3.5-turbo    & tie  & a & a\\
99  & vicuna-13b-v1.2  & gpt-3.5-turbo    & b & b & b\\
103 & gpt-3.5-turbo    & gpt-4            & b & b & b\\
104 & alpaca-13b       & gpt-3.5-turbo    & tie & tie & b\\
105 & gpt-3.5-turbo    & gpt-4            & b & tie & b\\
111 & llama-13b        & gpt-4            & b & b & b\\
112 & gpt-3.5-turbo    & claude-v1        & a  & tie & tie\\
115 & gpt-3.5-turbo    & claude-v1        & a & a & a\\
121 & vicuna-13b-v1.2  & gpt-4            & b & tie & b\\
124 & gpt-3.5-turbo    & gpt-4            & b & tie & tie\\
134 & gpt-3.5-turbo    & gpt-4            & b & b & b\\
140 & alpaca-13b       & gpt-3.5-turbo    & tie & b & b\\
141 & gpt-3.5-turbo    & gpt-4            & tie & tie & b\\
142 & gpt-3.5-turbo    & gpt-4            & b & b & b\\
143 & alpaca-13b       & vicuna-13b-v1.2  & tie & a & b\\
148 & gpt-3.5-turbo    & gpt-4            & b & a & b\\
148 & llama-13b        & claude-v1        & b & b & b\\
151 & llama-13b        & gpt-3.5-turbo    & b & b & b\\
152 & alpaca-13b       & vicuna-13b-v1.2  & a & tie & b\\
\bottomrule
\end{tabular}
\end{table*}

\section{The $M=2$ axioms and correlated experts}

Once one has determined the point in the QC and its corresponding product space
of $M=1$-consistent set of evaluations, we can consider the subset
of that space that is consistent with the $M=2$ set of axioms
for the $\binom{N}{2}$ pairs in an ensemble of $N$ classifiers. This subset
in single response space is a projection of the logically consistent
space that now must include variables of the form,
\begin{equation}
    R_{\elri{r}{i}, \elri{s}{j}}.
\end{equation}
These are the statistical summaries of how many times we observed
the two classifiers responding (\elri{r}{i}, \elri{s}{j}) on the
same question.

The unknown statistics of their aligned responses given true label
form $R$ copies of $R^2$ dimensional spaces. These are the counts
of how often the pair agreed and disagreed given true label. By
construction, they also lie on simplexes of one lower dimension
for each label. And, similar to the $M=1$ case, we need to
impose the inequalities,
\begin{equation}
    R_{\elri{r}{i}, \elri{s}{j}, \elt} \leq R_{\elri{r}{i}, \elri{s}{j}}.
\end{equation}
The count of any pair label event given true label cannot be higher
than the observed count of that event in their question aligned responses.

The $M=2$ axioms are also a set of $R$ linear equations but now in the
expanded space that includes the QC point, the $M=1$ product space point
for the pair, and the new pair response spaces given true label. We state
the axiom outright. For every \elt in the $R$ labels 
$(\eli{1}, \eli{1}, \ldots, \eli{R})$ and every pair of classifiers $i$
and $j$ the following linear relation is identically zero.
\begin{multline}
    R_{(\elt)_i, (\elt)_j; \elt} - Q_{\elt} \\
    + \sum_{c \in \{i, j\}} \sum_{\eli{r} \neq \elt} R_{\elri{r}{c}} 
    - \sum_{c \in \{i, j\}} \sum_{\eli{r} \neq \elt} \sum_{\eli{s} \neq \elt} R_{\elri{s}{c};\eli{r}} \\
    - \sum_{\eli{r} \neq \elt} R_{\elri{r}{i}, \elri{r}{j}}
    + \sum_{\eli{r} \neq \elt} \sum_{\eli{s} \neq \elt} R_{\elri{s}{i},\elri{s}{j}; \eli{r}} \\
    - \sum_{\eli{r} \neq \eli{s} \neq \elt} R_{\elri{r}{i},\elri{s}{j}; \elt}
\end{multline}

Its proof is exactly as the one for $M=1$ but now we
expand all response variables to pair responses given true label,
\begin{equation}
    R_{\elri{r}{i}, \elri{s}{j}, \elt}.
\end{equation}

Statistical measures of the correlation between pairs of 
classifiers can then be defined as,
\begin{equation}
    \Gamma_{\elri{r}{i}, \elri{s}{j}; \elt} = \frac{R_{\elri{r}{i}, \elri{s}{j}; \elt}}{Q_{\elt}} -
    \frac{R_{\elri{r}{i}; \elt}}{Q_{\elt}} \frac{R_{ \elri{s}{j};\elt}}{Q_{\elt}}
\end{equation}

\section{Classifiers and multiple-choice test takers have
the same unsupervised evaluation logic}

The lack of semantics in the logic explained here means that
its algebra applies equally well to classifiers or any
test taker responding to a Multiple Choice (MC) exam. The only
difference is how we interpret the response counts.

When we are doing classification, response `a' in one item/question
is pointing to the same thing as response `a' in another. In classification,
an evaluation of the classifiers is also telling us a statistic about the
items the classifiers labeled -- their prevalence.

In an MC exam, response letters need not have any semantic equality between
questions. Saying `b' in one question may mean something completely different
from answering `b' in another. The letters are a convenience that allows,
for example, students to fill out bubble-answer sheets. In the MC exam, the
ground truth for the QC point has no meaning outside the test. It is an artifact
of how the fixed, finite number of responses were encoded to letter responses.

The algebraic purpose of the label prevalences in an MC exam lies in that they
allow you to compute the percentage of test questions correct, $g_i$,
as,
\begin{equation}
    \label{eq:grade}
    g_i = \sum_{\eli{r} \in \mathcal{L}} \frac{Q_{\eli{r}}}{Q} \frac{R_{\elri{r}{i}, \eli{r}}}{Q_{\eli{r}}},
\end{equation}
for each classifier $i$ given QC point $(\qli{1},\ldots,\qli{R}).$ This defines
a set of logically consistent grades given observed test responses on an MC exam.

\section{If you solve the logic of single question summaries,
you solve the logic for question sequences}

Yet another advantage of logical consistency of counts of agreements/disagreements
being semantic free is that if you have axioms for the question-aligned summaries,
you immediately can write down the algebra for any sequence-aligned summaries
you may care to do.

Take the case of wanting to know statistics of correctness/errors for
consecutive pairs in a test. For a test of size $Q$ there are $Q-1$ such
pairs and we can map the $R^2$ tuples of the form $(\elri{r}{i}, \elri{s}{j})$
to $R^2$ labels. There is always an isomorphic mapping of whatever sequence of questions
we define for the test and $R^S$ labels where $S$ is the size of the sequences.
\end{document}